\documentclass[11pt,letterpaper]{article}
\usepackage{naaclhlt2016}
\usepackage{times}
\usepackage{latexsym}
\usepackage{graphicx}
\usepackage{multirow}
\usepackage{amsmath}
\usepackage{bbm}

\usepackage{makecell}
\usepackage{array}
\usepackage{pbox}

\usepackage{url}
\usepackage{color,soul}

\usepackage{footnote}

\renewcommand{\vec}[1]{\mathbf{#1}}

\naaclfinalcopy % Uncomment this line for the final submission
\def\naaclpaperid{680} %  Enter the naacl Paper ID here

\newcommand\B{\rule[-1.2ex]{0pt}{0pt}} % Bottom strut

\title{Joint Extraction of Events and Entities within a Document Context}

\author{Bishan Yang\\
	    Carnegie Mellon University\\
	    5000 Forbes Avenue\\
	    Pittsburgh, PA, 15213\\
	    {\tt bishan@cs.cmu.edu}
	  \And
	Tom Mitchell\\
  	Carnegie Mellon University\\
  	5000 Forbes Avenue\\
  	Pittsburgh, PA, 15213\\
  {\tt tom.mitchell@cs.cmu.edu}}

\date{}

\begin{document}

\maketitle

\begin{abstract}
Events and entities are closely related; entities are often actors or participants in events and events without entities are uncommon. The interpretation of events and entities is highly contextually dependent. Existing work in information extraction typically models events separately from entities, and performs inference at the sentence level, ignoring the rest of the document. In this paper, we propose a novel approach that models the dependencies among variables of events, entities, and their relations, and performs joint inference of these variables across a document. The goal is to enable access to document-level contextual information and facilitate context-aware predictions. We demonstrate that our approach substantially outperforms the state-of-the-art methods for event extraction as well as a strong baseline for entity extraction.
\end{abstract}

\section{Introduction}
Events are things that happen or occur; they involve entities (people, objects, etc.) who perform or are affected by the events and spatio-temporal aspects of the world. Understanding events and their descriptions in text is necessary for any generally-applicable machine reading systems. It is also essential in facilitating practical applications such as news summarization, information retrieval, and knowledge base construction.

The interpretation of event descriptions is highly contextually dependent. To make correct predictions, a model needs to account for mentions of events and entities together with the discourse context. Consider, for example, the following excerpt from a news report:
\begin{quotation}
``On \textit{Thursday}, there was a massive \textit{U.S.} \textbf{aerial bombardment} in which more than 300 \textit{Tomahawk cruise missiles} rained down on \textit{Baghdad}. \textit{Earlier Saturday}, \textit{Baghdad} was again \textbf{targeted}. ...''
\end{quotation}
The excerpt describes two U.S. attacks on Baghdad. The two event anchors (triggers) are boldfaced and the mentions of entities and spatio-temporal information are italicized. The first event anchor ``aerial bombardment'' along with its surrounding entity mentions --- ``U.S.'', ``Tomahawk cruise missiles'', and ``Baghdad'', describe an attack from the U.S. on Baghdad with Tomahawk cruise missiles being the weapon. The second sentence on its own contains little event-related information, but together with the context of the previous sentence, it indicates another U.S. attack on Baghdad.

State-of-the-art event extraction systems have difficulties inferring such information due to two main reasons. First, they extract events and entities in separate stages: entities such as people, organization, and locations are first extracted by a named entity tagger, and then these extracted entities are used as inputs for extracting events and their arguments~\cite{li2013joint}. This often causes errors to propagate. In the above example, if the entity tagger mistakenly identifies ``Baghdad'' as a person, then the event extractor will fail to extract ``Baghdad'' as the place where the attack happened. In fact, previous work~\cite{li2013joint} observes that using previously extracted entities in event extraction results in a substantial decrease in performance compared to using gold-standard entity information.

Second, most existing work extracts events independently from each individual sentence, ignoring the rest of the document~\cite{li2013joint,judea2015event,nguyen2015event}. Very few attempts have been made to incorporate document context for event extraction. Ji and Grishman~\shortcite{ji2008refining} model the information flow in two stages: the first stage trains classifiers for event triggers and arguments within each sentence; the second stage applies heuristic rules to adjust the classifiers' outputs to satisfy document-wide (or document-cluster-wide) consistency. Liao and Grishman~\shortcite{liao2010using} further improved the rule-based inference by training additional classifiers for event triggers and arguments using document-level information. Both approaches only propagate the highly confident predictions from the first stage to the second stage. To the best of our knowledge, there is no unified model that jointly extracts events from sentences across the whole document.

In this paper, we propose a novel approach that simultaneously extracts events and entities within a document context.\footnote{The code for our system is available at \url{https://github.com/bishanyang/EventEntityExtractor}.} We first decompose the learning problem into three tractable subproblems: (1) learning the dependencies between a single event and all of its potential arguments, (2) learning the co-occurrence relations between events across the document, and (3) learning for entity extraction. Then we combine the learned models for these subproblems into a joint optimization framework that simultaneously extracts events, semantic roles, and entities in a document. In summary, our main contributions are:
\begin{enumerate}
\item We propose a structured model for learning within-event structures that can effectively capture the dependencies between an event and its arguments, and between the semantic roles and entity types for the arguments.
\item We introduce a joint inference framework that combines probabilistic models of within-event structures, event-event relations, and entity extraction for joint extraction of the set of entities and events over the whole document.
\item We conduct extensive experiments on the Automatic Content Extraction (ACE) corpus, and show that our approach significantly outperforms the state-of-the-art methods for event extraction and a strong baseline for entity extraction.  
\end{enumerate}

\section{Task Definition}
\label{sec:taskdef}
We adopt the ACE definition for entities~\cite{ACEentity} and events~\cite{ACEevent}:
\begin{itemize}
\item \textbf{Entity mention}: An entity is an object or set of objects in the world. An entity mention is a reference to an entity in the form of a noun phrase or a pronoun.
\item \textbf{Event trigger}: the word or phrase that clearly expresses its occurrence. Event triggers can be verbs, nouns, and occasionally adjectives like ``dead'' or ``bankrupt''.
\item \textbf{Event argument}: event arguments are entities that fill specific roles in the event. They mainly include participants (i.e., the entities that are involved in the event) and general event attributes such as place and time, and some event-type-specific attributes that have certain values (e.g., {\sc Job-Title}, {\sc Crime}).
\end{itemize}

We are interested in extracting entity mentions, event triggers, and event arguments. We consider ACE entity types {\sc PER}, {\sc ORG}, {\sc GPE}, {\sc LOC}, {\sc FAC}, {\sc VEH}, {\sc WEA} and ACE {{\sc Value} and {{\sc Time} expressions\footnote{To simplify notation, we include values and times when referring to entities in the rest of the paper.}, and focus on 33 ACE event subtypes, each of which has its own set of semantic roles for the potential arguments. There are 35 such roles in total, but we collapse 8 of them that are time-related (e.g., {{\sc Time-Holds}, {{\sc Time-At-End}) into one, because most of these roles have very few training examples. Figure~\ref{anno_example} shows an example of ACE annotations for events and entities in a sentence. Note that not every entity mention in the sentence is involved in events and a single entity mention can be associated with multiple events. 

\begin{figure*}
  \centering
    \includegraphics[width=1.0\textwidth]{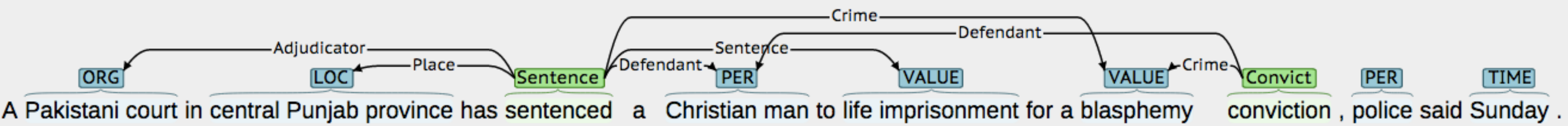}
    \label{anno_example}
    \caption{An example of ACE annotations of events and entities. The event triggers and the entity mentions are marked in different colors. Each event trigger has an event subtype marked above it and each entity mention has an entity type marked above it. Each event trigger evokes an event with semantic roles that are filled by entity mentions. The roles are marked on the links between event trigger and entity mentions. For example, ``conviction'' evokes a {\sc Convict} event, and has the {\sc Crime} and {\sc Defendant} roles filled by ``blasphemy'' and ``Christian man'' respectively. }
\end{figure*}

\section{Approach}
In this section, we describe our approach for joint extraction of events and entities within a document context. We first decompose the learning problem into three tractable subproblems: learning within-event structures, learning event-event relations, and learning for entity extraction. We will describe the probabilistic models for learning these subproblems. Then we present a joint inference framework that integrates these learned models into a single model to jointly extract events and entities across a document.  

\subsection{Learning Within-event Structures}
\label{within-event-model}
As described in Section~\ref{sec:taskdef}, a mention of an event consists of an event trigger and a set of event arguments. Each event argument is also an entity mention with an entity type. In the following, we develop a probabilistic model to learn such dependency structure for each individual event mention.

Given a document $x$, we first generate a set of event trigger candidates $\mathcal{T}$ and a set of entity candidates $\mathcal{N}$.\footnote{We describe how to extract these candidates in Section~\ref{experiment}.} For each trigger candidate $i\in \mathcal{T}$, we associate it with a discrete variable $t_i$ that takes values from the 33 ACE event types and a {\sc None} class indicating other events or no events. Denote the set of entity candidates that are potential arguments for trigger candidate $i$ as $\mathcal{N}_i$.\footnote{In this paper we only consider entity mentions that are in the same sentence as the trigger to be potential event arguments due to the ACE annotations. However, our model is general and can handle event-argument relations across sentences with appropriate features.} For each $j\in \mathcal{N}_i$, we associate it with a discrete variable $r_{ij}$ which models the event-argument relation between trigger candidate $i$ and entity candidate $j$. It takes values from 28 semantic roles and a {\sc None} class indicating invalid roles. Each argument candidate $j$ is also associated with an entity type variable $a_{j}$, which takes values from 9 entity types and a {\sc None} class indicating invalid entity types. 
\begin{figure}
\centering
\includegraphics[width=0.5\linewidth]{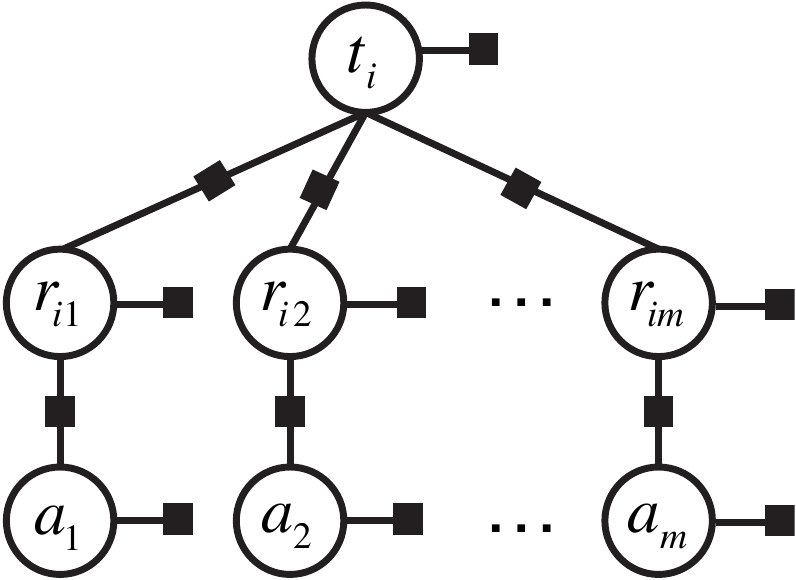}
\caption{A factor graph representation of the within-event model, relating the event type $t_i$ of trigger candidate $i$ to the role type $r_{ij}$ of each argument candidate $j$ along with its entity type $a_j$.}
\label{fig:model}
\end{figure}

We define the joint distribution of variables $t_i$, $\vec{r}_{i\cdot}=\{r_{ij}\}_{j\in \mathcal{N}_i}$, and $\vec{a}_{\cdot}=\{a_{j}\}_{j\in \mathcal{N}_i}$ conditioned on the observations, which can be factorized according to the factor graph shown in Figure~\ref{fig:model}:
\begin{equation}
\begin{split}
&p_{\boldsymbol\theta}(t_i, \vec{r}_{i\cdot}, \vec{a}_{\cdot}|i,\mathcal{N}_i,x)\propto\exp\bigg(\boldsymbol\theta_1^Tf_1(t_i,i,x)+\\
&\sum_{j\in \mathcal{N}_i}\boldsymbol\theta_2^Tf_2(r_{ij},i,j,x)+\sum_{j\in \mathcal{N}_i}\boldsymbol\theta_3^Tf_3(t_i,r_{ij},i,j,x)+\\
&\sum_{j\in \mathcal{N}_i}\boldsymbol\theta_4^Tf_4(a_{j},j,x)+\sum_{j\in \mathcal{N}_i}\boldsymbol\theta_5^Tf_5(r_{ij},a_{j},j,x)\bigg)
\end{split}
\end{equation}
where $\boldsymbol\theta_1,...,\boldsymbol\theta_5$ are vectors of parameters that need to be estimated, and $f_1,...,f_5$ are different forms of feature functions which we will describe later. 

Note that not all configurations of the variables are valid in our model. Based on the definitions in Section~\ref{sec:taskdef}, each event type takes arguments with certain semantic roles. For example, the arguments of the event {\sc Marry} can only play the roles of {\sc Person}, {\sc Time}, and {\sc Place}. In addition, a {\sc None} event type should not take any arguments. Similarly, each semantic role should be filled with entities with compatible types. For example, the {\sc Person} role type can only be filled with an entity of type {\sc PER}. However, a {\sc None} role type can be filled with an entity of any type. To account for these compatibility constraints, we enforce the probabilities of all invalid configurations to be zero. 

\textbf{Features}. $f_1$, $f_2$, and $f_4$ are unary feature functions that depend on trigger variable $t_i$, argument variable $r_{ij}$, and entity variable $a_j$ respectively. We construct a set of features for each feature function (see Table~\ref{table:features}). Many of these features overlap with those used in previous work~\cite{li2013joint,li2014constructing}, except for the word embedding features for triggers and the features for entities which are derived from multiple entity resources. $f_3$ and $f_5$ are pairwise feature functions that depend on trigger-argument pair $(t_i, r_{ij})$ and argument-entity pair $(r_{ij}, a_j)$ respectively. We consider simple indicator functions $\mathbbm{1}_{t,r}$ and $\mathbbm{1}_{r,a}$ as features ($\mathbbm{1}_y(x)$ equals 1 when $x=y$ and $0$ otherwise).

\begin{table*}[t]
\centering
\begin{footnotesize}
\begin{tabular}{|c|c|c|}
\hline
Category & Type & Features  \\ \hline
\multirow{8}{*}{Trigger} & \multirow{6}{*}{\parbox{3cm}{\textit{Lexical resources:}\\
WordNet\\
Nomlex\\
FrameNet\\
Word2Vec}} & \parbox{10cm}{1. lemmas of the words in the trigger mention}\\
& & \parbox{10cm}{2. nominalization of the words based on Nomlex~\cite{macleod1998nomlex}}\\
& & \parbox{10cm}{3. context words within a window of size $2$}\\
& & \parbox{10cm}{4. similarity features between the head word and a list of trigger seeds based on WordNet~\cite{bronstein2seed}}\\
& & \parbox{10cm}{5. semantic frames that associate with the head word and its p-o-s tag based on FrameNet~\cite{li2014constructing}}\\
& &\parbox{10cm}{6. pre-trained vector for the head word~\cite{mikolov2013distributed}}\\
\cline{2-3}
\rule{0pt}{4ex}  
& \multirow{2}{*}{\parbox{3cm}{\textit{Syntactic resources}:\\
Stanford parser}} &\parbox{10cm}{7. dependency edges involving the head word, both lexicalized and unlexicalized}\\ 
& &\parbox{10cm}{8. whether the head word is a pronoun}\\
%& &\parbox{10cm}{}\\
\hline

\multirow{6}{*}{Argument} & \multirow{3}{*}{\parbox{3cm}{\textit{Lexical resources:}\\
WordNet}} & \parbox{10cm}{1. lemmas of the words in the entity mention}\\
& &\parbox{10cm}{2. lemmas of the words in the trigger mention}\\
& &\parbox{10cm}{3. words between the entity mention and the trigger mention}\\
\cline{2-3}
\rule{0pt}{4ex}  
& \multirow{3}{*}{\parbox{3cm}{\textit{Syntactic resources}:\\
Stanford parser}} &\parbox{10cm}{4. the relative position of the entity mention to the trigger mention (before, after, or contain)}\\ 
& &\parbox{10cm}{5. whether the entity mention and the trigger mention are in the same clause}\\
& &\parbox{10cm}{6. the shortest dependency paths between the entity mention and the trigger mention\B}\\
\hline

\multirow{4}{*}{Entity} & \multirow{4}{*}{\parbox{3cm}{\textit{Entity resources:}\\
Stanford NER\\
NELL KB}} & \parbox{10cm}{1. Gender and animacy attributes of the entity mention}\\
& & \parbox{10cm}{2. Stanford NER type for the entity mention}\\
& & \parbox{10cm}{3. Semantic type for the entity mention based on the NELL knowledge base~\cite{NELL-aaai15}}\\
& & \parbox{10cm}{4. Predicted entity type and confidence score for the entity mention output by the entity extractor described in Section~\ref{entity_extractor}\B}\\
\hline
\end{tabular}
\caption{Features for event triggers, event arguments, and entity mentions.}
\label{table:features}
\end{footnotesize}
\end{table*}

\textbf{Training.} For model training, we find the optimal parameters $\boldsymbol\theta$ using the maximum-likelihood estimates with an L2 regularization:
$$\boldsymbol\theta^*=\arg\max_{\boldsymbol\theta}\mathcal{L}(\boldsymbol\theta)-\lambda||\boldsymbol\theta||_2^2$$
$$\mathcal{L}(\boldsymbol\theta)=\sum_i\log p(t_i, \vec{r}_{i\cdot}, \vec{a}_{\cdot}|i,\mathcal{N}_i,x)$$ We use L-BFGS to optimize the training objective. To calculate the gradient, we use the sum-product algorithm to compute the exact marginals for the unary cliques $t_i$, $r_{ij}$, $a_{j}$ and the pairwise cliques $(t_i,r_{ij}), (r_{ij},a_{j})$. Typically the training complexity for graphical models with unary and pairwise cliques is quadratic in the size of the label set. However, the complexity of our model is much lower than that since we only need to compute the joint distributions over valid variable configurations. Denote the number of event subtypes as $T$, the number of event argument roles as $N$, the average number of argument roles for each event subtype as $k_1$, the average number of entity types for each event argument as $k_2$, and the average number of argument candidates for each trigger candidate as $M$. The complexity of computing the joint distribution is $O(M\times(k_1T+k_2N))$, and $k_1$ and $k_2$ are expected to be small in practice ($k_1=6,k_2=3$ in ACE).

\subsection{Learning Event-Event Relations}
\label{event-pair}
So far we have described a model for learning structures for a single event. However, the inference of the event types for individual events may depend on other events that are mentioned in the document. For example, an {\sc Attack} event is more likely to occur with {\sc Injure} and {\sc Die} events than with life events like {\sc Marry} and {\sc Born}. In order to capture this intuition, we develop a pairwise model of event-event relations in a document. 

Our training data consists of all pairs of trigger candidates that co-occur in the same sentence or are connected by a coreferent subject/object if they are in different sentences.\footnote{We use the Stanford coreference system~\cite{lee2013deterministic} for within-document entity coreference.} We want to propagate information between these trigger pairs since they are more likely to be related.

Formally, given a trigger candidate pair $(i,i')$, we estimate the probabilities for their event types $(t_i,t_{i'})$ as
\begin{equation}
p_{\boldsymbol\phi}(t_i, t_{i'}|x,i,i')\propto\exp\Big(\boldsymbol\phi^Tg(t_{i}, t_{i'},x,i,i')\Big)
\end{equation}
where $\boldsymbol\phi$ is a vector of parameters and $g$ is a feature function that depends on the trigger candidate pair and their context. We consider both trigger-specific features and relational features. For trigger-specific features, we use the same trigger features listed in Table~\ref{table:features}. For relational features, we consider for each pair of trigger candidates: (1) whether they are connected by a conjunction dependency relation (based on dependency parsing); (2) whether they share a subject or an object (based on dependency parsing and coreference resolution); (3) whether they have the same head word lemma; (4) whether they share a semantic frame based on FrameNet. During training, we use L-BFGS to compute the maximum-likelihood estimates of $\phi$.

\subsection{Entity Extraction}
\label{entity_extractor}
For entity extraction, we trained a standard linear-chain Conditional Random Field (CRF)~\cite{lafferty2001conditional} using the BIO scheme (i.e., identifying the \textbf{B}eginning, the \textbf{I}nside and the \textbf{O}utside of the text segments). We use features that are similar to those from previous work~\cite{ratinov2009design}: (1) current words and part-of-speech tags; (2) context words in a window of size 2; (3) word type such as all-capitalized, is-capitalized, and all-digits; (4) Gazetteer-based entity type if the current word matches an entry in the gazetteers collected from Wikipedia~\cite{ratinov2009design}. In addition, we consider pre-trained word embeddings~\cite{mikolov2013distributed} as dense features for each word in order to improve the generalizability of the model.

\subsection{Joint Inference}
\label{joint-inference}
Our end goal is to extract coherent event mentions and entity mentions across a document. To achieve this, we propose a joint inference approach that allows information flow among the three local models and finds globally-optimal assignments of all variables, including the trigger variables $\vec{t}$, the argument role variables $\vec{r}$, and the entity variables $\vec{a}$. Specifically, we define the following objective:
\begin{equation}
\max_{\vec{t},\vec{r},\vec{a}}\sum_{i\in \mathcal{T}}E(t_i,\vec{r}_{i\cdot},\vec{a}_{\cdot})+\sum_{i,i'\in \mathcal{T}}R(t_{i},t_{i'})+\sum_{j\in \mathcal{N}}D(a_j)
\end{equation}
The first term is the sum of confidence scores for individual event mentions based on the parameter estimates from the within-event model. $E(t_i,\vec{r}_{i\cdot},\vec{a}_{\cdot})$ can be further decomposed into three parts. \begin{equation*}
\begin{split}
&E(t_i,\vec{r}_{i\cdot},\vec{a}_{\cdot})=\\
&\log p_{\boldsymbol\theta}(t_i|i,\mathcal{N}_i,x)+\sum_{j\in \mathcal{N}_i}\log p_{\boldsymbol\theta}(t_i,r_{ij}|i,\mathcal{N}_i,x)\\
&+\sum_{j\in \mathcal{N}_i}\log p_{\boldsymbol\theta}(r_{ij},a_j|i,\mathcal{N}_i,x)\\
\end{split}
\end{equation*} The second term is the sum of confidence scores for event relations based on the parameter estimates from the pairwise event model, where $R(t_{i},t_{i'})=\log p_{\boldsymbol\phi}(t_i, t_{i'}|i,i',x)$. The third term is the sum of confidence scores for entity mentions, where $D(a_j)=\log p_{\psi}(a_j|j,x)$ and $p_{\psi}(a_j|j,x)$ is the marginal probability derived from the linear-chain CRF described in Section~\ref{entity_extractor}. The optimization is subjected to agreement constraints that enforce the overlapping variables among the three components to agree on their values. 

The joint inference problem can be formulated as an integer linear program (ILP). To solve it efficiently, we find solutions for the relaxation of the problem using a dual decomposition algorithm AD$^3$~\cite{martins2011augmented}. AD$^3$ has been shown to be orders of magnitude faster than a general purpose ILP solver in practice~\cite{das2012exact}. It is also particularly suitable for our problem since it involves decompositions that have many overlapping simple factors. We observed that AD$^3$ recovers the exact solutions for all the test documents in our experiments and the runtime for labeling each document is only three seconds in average in a 64-bit machine with two 2GHz CPUs and 8GB of RAM.

\section{Experiments}
\label{experiment}
We conduct experiments on the ACE2005 corpus.\footnote{\url{http://www.itl.nist.gov/iad/mig/tests/ace/2005/}} It contains text documents from a variety of sources such as newswire reports, weblogs, and discussion forums. We use the same data split as in Li et al.~\shortcite{li2013joint}. Table~\ref{ace_data} shows the data statistics.

We adopt the evaluation metrics for events as defined in Li et al.~\shortcite{li2013joint}. An event trigger is correctly identified if its offsets match those of a gold-standard trigger; and it is correctly classified if its event subtype (33 in total) also match the subtype of the gold-standard trigger. An event argument is correctly identified if its offsets and event subtype match those of any of the reference argument mentions in the document; and it is correctly classified if its semantic role (28 in total) is also correct. For entities, a predicted mention is correctly extracted if its head offsets and entity type (9 in total) match those of the reference entity mention. 

Note that our approach requires entity mention candidates and event trigger candidates as input. Instead of enumerating all possible text spans, we generate high-quality entity mentions from the $k$-best predictions of our CRF entity extractor (in Section~\ref{entity_extractor}).\footnote{During training, we randomly split the training data into 10 parts and consider the $k$-best predictions for each part.} Similarly, we train a CRF for event trigger extraction using the same features except for the gazetteers, and generate trigger candidates based on the k-best predictions. We set $k=50$ for entities and $k=10$ for event triggers based on performance on the development set. They cover 92.3\% of the gold-standard entity mentions and 96.3\% of the gold-standard event triggers in the test set.
 
\begin{table}
\begin{footnotesize}
\begin{center}
\begin{tabular}{|c|c|c|c|}
\hline
& Train & Dev & Test\\
\hline
Documents & 529 & 40 & 30 \\
\hline
Sentences & 14,837 & 863 & 672\\
\hline
Triggers & 4,337 & 497 & 438\\
\hline
Arguments & 7,768 & 933 & 911\\
\hline
Entity Mentions & 48,797 & 3,917 & 4,184\\
\hline
\end{tabular}
\end{center}
\caption{\label{ace_data}Statistics of the ACE2005 dataset.}
\end{footnotesize}
\end{table}

\subsection{Results}
\makesavenoteenv{table}
\begin{table*}[t]
\begin{footnotesize}
\begin{center}
\begin{tabular}{|c|c|c|c|c|c|c|c|c|c|c|c|c|}
\hline
& \multicolumn{3}{c|}{\makecell{Event Trigger\\Identification}} & \multicolumn{3}{c|}{\makecell{Event Trigger\\Classification}} & \multicolumn{3}{c|}{\makecell{Event Argument\\Identification}} & \multicolumn{3}{c|}{\makecell{Argument Role\\Classification}}\\
Model & P & R & F1 & P & R & F1 & P & R & F1 & P & R & F1\\
\hline
{\sc JointBeam}~\cite{li2013joint} & 76.6 & 58.7 & 66.5 & 74.0 & 56.7 & 64.2 & 74.6 & 25.5 & 38.0 & 68.8 & 23.5 & 35.0\\
\hline
{\sc StagedMaxEnt} & 73.9 & \bf{66.5} & 70.0 & 70.4 & 63.3 & 66.7 & \bf{75.7} & 20.2 & 31.9 & \bf{71.2} & 19.0 & 30.0\\
\hline
{\sc WithinEvent} & 76.9 & 63.8 & 69.7 & 74.7 & 62.0 & 67.7 & 72.4 & 37.2 & 49.2 & 69.9 & 35.9 & 47.4 \\
\hline
{\sc JointEventEntity} & \bf{77.6} & 65.4 & \bf{71.0}$^{\ast}$ & \bf{75.1} & 63.3 & \bf{68.7} & 73.7 & \bf{38.5} & \bf{50.6}$^{\ast}$ & 70.6 & \bf{36.9} & \bf{48.4}$^{\ast}$\\
\hline
\end{tabular}
\end{center}
\caption{\label{event_results}Event extraction results on the ACE2005 test set. $\ast$ indicates that the difference in F1 compared to the second best model ({\sc WithinEvent}) is statistically significant ($p<0.05$).}
\end{footnotesize}
\end{table*}
%where $p$ is estimated by the bootstrap procedure~\cite{berg2012empirical} with 10,000 samples of the test documents

\textbf{Event Extraction}. We compare the proposed models {\sc WithinEvent} (in Section~\ref{within-event-model}) and {\sc JointEventEntity} (in Section~\ref{joint-inference}) with two strong baselines. One is {\sc JointBeam}~\cite{li2013joint}, a state-of-the-art event extractor that uses a structured perceptron with beam search for sentence-level joint extraction of event triggers and arguments. The other is {\sc StagedMaxEnt}, a typical two-stage approach that detects event triggers first and then event arguments. We use the same event trigger candidates and entity mention candidates as input to all the comparing models except for {\sc JointBeam}, because {\sc JointBeam} only extracts event mentions and assumes entity mentions are given. We consider a realistic experimental setting where no gold-standard annotations are available for entities during testing. To obtain results from {\sc JointBeam}, we ran the actual system\footnote{\url{https://github.com/oferbr/BIU-RPI-Event-Extraction-Project}} used in Li et al.~\shortcite{li2013joint} using the entity mentions output by our CRF-based entity extractor. 

Table~\ref{event_results} shows the average\footnote{We report the micro-average scores as in previous work~\cite{li2013joint}.} precision, recall, and F1 score for event triggers and event arguments. We can see that our {\sc WithinEvent} model, which explicitly models the trigger-argument dependencies and argument-role-entity-type dependencies, outperforms the MaxEnt pipeline, especially in event argument extraction. This shows that modeling the trigger-argument dependencies is effective in reducing error propagation. 

Comparing to the state-of-the-art event extractor {\sc JointBeam}, the improvements introduced by {\sc WithinEvent} are substantial in both event triggers and event arguments. We believe there are two main reasons: (1) {\sc WithinEvent} considers all possible joint trigger/argument label assignments, whereas {\sc JointBeam} considers only a subset of the possible assignments based on a heuristic beam search. More specifically, when predicting labels for token $i$, JointBeam considers only the K-best ($K=4$ in their paper) partial trigger/argument label configurations for the previous $i-1$ tokens. As the length of the sentence increases, a large amount of information will be thrown away. (2) {\sc WithinEvent} models argument-role-entity-type dependencies, whereas {\sc JointBeam} assumes the entity types are given. This can cause error propagation.

{\sc JointEventEntity} provides the best performance among all the models on all evaluation categories. It boosts both precision and recall compared to {\sc WithinEvent}.\footnote{All significance tests reported in this paper were computed
using the paired bootstrap procedure~\cite{berg2012empirical} with 10,000 samples of the test documents.} This demonstrates the advantages of {\sc JointEventEntity} in allowing information propagation across event mentions and entity mentions and making more context-aware and semantically coherent predictions. 

\begin{table}
\begin{footnotesize}
\begin{center}
\begin{tabular}{|c|c|c|}
\hline
Model & Trigger & Arg\\
\hline
{\sc Cross-doc}~\cite{ji2008refining} & 67.3 & 42.6\\
\hline
{\sc CNN}~\cite{nguyen2015event} & 67.6 & -\\
\hline
{\sc JointEventEntity} & \bf{68.7} & \bf{48.4}\\
\hline
\end{tabular}
\end{center}
\caption{\label{event_f1_results}Comparison of the results (F1 score) of {\sc JointEventEntity} and the best known results on ACE event trigger classification and argument role classification.}
\end{footnotesize}
\end{table}

\begin{table}
\begin{footnotesize}
\begin{center}
\begin{tabular}{|c|c|c|c|}
\hline
Model & P & R & F1\\
\hline
{\sc CRFEntity} & \bf{85.5} & 73.5 & 79.1\\
\hline
{\sc JointEventEntity} & 82.4 & \bf{79.2} & \bf{80.7}$^{\ast}$\\
\hline
\end{tabular}
\end{center}
\caption{\label{ace_entity_results}Entity extraction results on the ACE2005 test set. $\ast$ indicates statistical significance ($p<0.05$).}
\end{footnotesize}
\end{table}

We also compare the results of {\sc JointEventEntity} with the best known results on the ACE event extraction task in Table~\ref{event_f1_results}. {\sc Cross-doc}~\cite{ji2008refining} performs cross-document inference of events using document clustering information, and {\sc CNN}~\cite{nguyen2015event} is a convolutional neural network for extracting event triggers at the sentence level. We see that {\sc JointEventEntity} outperforms both models and achieves new state-of-the-art results for event trigger and argument extraction in an end-to-end evaluation setting. 

\begin{table}
\begin{footnotesize}
\begin{center}
\begin{tabular}{|c|c|c|c|c|}
\hline
Model & {\sc PER} & {\sc GPE} & {\sc ORG} & {\sc TIME}\\
\hline
{\sc CRFEntity} & 85.1 & 87.0 & 65.4 & 78.4\\
\hline
{\sc JointEventEntity} & \bf{87.1} & 87.0 & \bf{70.2} & \bf{80.2}\\
\hline
\end{tabular}
\end{center}
\caption{\label{entity_detail_results}Entity extraction results (F1 score) per entity type.}
\end{footnotesize}
\end{table}

\textbf{Entity Extraction}. In addition to extracting event mentions, {\sc JointEventEntity} also extracts entity mentions. We compare its output with the output of a strong entity extraction baseline {\sc CRFEntity} (described in Section~\ref{entity_extractor}). Table~\ref{ace_entity_results} shows the (micro-)average precision, recall, and F1 score. We see that {\sc JointEventEntity} introduces a significant improvement in recall and F1. Table~\ref{entity_detail_results} further shows the F1 score for four major entity types {\sc PER}, {\sc GPE}, {\sc ORG}, and {\sc TIME} in ACE. The promising improvements indicate that joint modeling of events and entities allows for more accurate predictions about not only events but also entities.

\subsection{Error Analysis}
Table~\ref{error_analysis} divides the errors made by {\sc JointEventEntity} based on different subtasks and the classification error types in each task. For event triggers, the majority of the errors relates to missing triggers and only 3.7\% involves misclassified event types (e.g., a {\sc Demonstration} event is mistaken for a {\sc Transport} event). Among the missing triggers, we examine the cases where the event types are correctly identified in a sentence but with incorrect triggers and find that there are only 5\% of such cases. For event arguments, the majority of the errors relates to missing arguments and only 4.1\% is about misclassified argument roles. Among the missing event arguments, 10\% of them has correctly identified entity types. 

\begin{table}
\begin{footnotesize}
\begin{center}
\begin{tabular}{|c|c|c|c|}
\hline
\makecell{Error Type}  & Missing & Spurious & Misclassified\\
\hline
{\sc Trigger} & 62.1\% & 34.2\% & 3.7\% \\
\hline
{\sc Argument} & 71.2\% & 24.7\% & 4.1\% \\
\hline
{\sc Entity} & 43.4\% & 30.5\% & 26.1\% \\
\hline
\end{tabular}
\end{center}
\caption{\label{error_analysis} Classification of errors made by {\sc JointEventEntity}.}
\end{footnotesize}
\end{table}

In general, the errors for event extraction are commonly due to three reasons: (1) Lexical sparsity. For example, in the sentence ``At least three members of a family ... were \textbf{hacked} to death ...'', our model fails to detect that ``hacked'' triggers an {\sc Attack} event, because it has never seen ``hacked'' with this sense during training. Using WordNet and pre-trained word vectors may alleviate the sparsity issue. It is also important to disambiguate word senses in context. (2) Shallow understanding of context, especially long-range context. For example, given the sentence ``\textbf{She} is being held on 50,000 dollars bail on a charge of first-degree reckless \textbf{homicide} ...'', the model detects that ``homicide'' triggers an event, but fails to detect that ``She'' refers to the {\sc Agent} who committed the homicide. This is mainly due to the complex long-distance dependency between the trigger and the argument. (3) Use of complex language such as metaphor, idioms, and sarcasm. Addressing these phenomena is in general difficult since it requires richer background knowledge and more sophisticated inference. 

For entity extraction, we find that integrating event information into entity extraction successfully improves recall and F1. However, since the ACE dataset is restricted to a limited set of events, a large portion of the sentences does not contain any event triggers and event arguments that are of interest. For these sentences, there is little or no benefit of joint modeling. We also find that some entity misclassification errors can be avoided if entity coreference information is available. We plan to investigate coreference resolution as an additional component to our joint model in future work. 

\section{Related Work}
Event extraction has been mainly studied using the ACE data~\cite{doddington2004automatic} and biomedical data for the BioNLP shared tasks~\cite{kim2009overview}. To reduce task complexity, early work employs a pipeline of classifiers that extracts event triggers first, and then determines their arguments~\cite{ahn2006stages,bjorne2009extracting}. Recently, Convolutional Neural Networks have been used to improve the pipeline classifiers~\cite{nguyen2015event,chen2015event}. As pipeline approaches suffer from error propagation, researchers have proposed methods for joint extraction of event triggers and arguments, using either structured perceptron~\cite{li2013joint}, Markov Logic~\cite{poon2010joint}, or dependency parsing algorithms~\cite{mcclosky2011event}. However, existing joint models largely rely on heuristic search to aggressively shrink the search space. One exception is work in Riedel and McCallum~\shortcite{riedel2011fast}, which uses dual decomposition to solve joint inference with runtime guarantees. Our work is similar to Riedel and McCallum~\shortcite{riedel2011fast}. However, there are two main differences: first, our model extracts both event mentions and entity mentions; second, it performs joint inference across sentence boundaries. Although our approach is evaluated on ACE, it can be easily adapted to BioNLP data by using appropriate features for events triggers, argument roles, and entities. We consider this as future work. 

There has been work on improving event extraction by exploiting document-level context.~\newcite{berant2014modeling} exploits event-event relations, e.g., causality, inhibition, which frequently occur in biological texts. For general texts most work focuses on exploiting temporal event relations~\cite{chambers2008jointly,do2012joint,mcclosky2012learning}. For the ACE domain, there is work on utilizing event type co-occurrence patterns to propagation event classification decisions~\cite{ji2008refining,liao2010using}. Our model is similar to their work. It models the co-occurrence relations between event types (e.g., a {\sc Die} event tends to co-occur with {\sc Attack} events and {\sc Transport} events). It can be extended to handle other types of event relations (e.g., causal and temporal) by designing appropriate features. Chambers and Jurafsky~\shortcite{chambers2009unsupervised,chambers2011template} learn narrative schemas by linking event verbs that have coreferring syntactic arguments. Our model also adopts this intuition to relate event triggers across sentences. In addition, each event argument is grounded by its entity type (e.g., an entity mention of type {\sc PER} can only fill roles that can be played by a person). 

%Existing work in information extraction largely treats entity extraction as a separate task and uses its predictions as input to higher level information extraction tasks such as event extraction. This results in cascading errors and low end-to-end performance~\cite{li2013joint}. Unlike previous work, our work treats entities as hidden variables and performs joint inference over variables of events and entities.

\section{Conclusion}
In this paper, we introduce a new approach for automatic extraction of events and entities across a document. We first decompose the learning problem into three tractable subproblems: learning within-event structures, learning event-event relations, and learning for entity extraction. We then integrate these learned models into a single model that performs joint inference of all event triggers, semantic roles for events, and entities across the whole document. Experimental results demonstrate that our approach outperforms the state-of-the-art event extractors by a large margin and substantially improves a strong entity extraction baseline. For future work, we plan to integrate entity and event coreference as additional components into the joint inference framework. We are also interested in investigating the integration of more sophisticated event-event relation models of causality and temporal ordering. 

\section*{Acknowledgments}
This work was supported in part by NSF grant IIS-1250956, and in part by the DARPA DEFT program under contract FA87501320005. We would like to thank members of the CMU NELL group for helpful comments. We also thank the anonymous reviewers for insightful suggestions.

\bibliography{naaclhlt2016}
\bibliographystyle{naaclhlt2016}

\end{document}